\documentclass[letterpaper, 10 pt, conference]{ieeeconf}  

\IEEEoverridecommandlockouts                              
                                                          
\usepackage{graphicx}
\usepackage{times}
\usepackage{amsmath}
\usepackage{amssymb}
\usepackage{multicol}
\usepackage{float}
\usepackage{subcaption}
\usepackage{color}
\usepackage{bbding}
\usepackage{balance}

\definecolor{markupcolour}{rgb}{0.7,0.0,0.1}

\usepackage[pagebackref=false,breaklinks=true,colorlinks,bookmarks=false]{hyperref} %

\title{EasyLabel: A Semi-Automatic Pixel-wise Object Annotation Tool \\ for Creating Robotic RGB-D Datasets}
\author{
Markus Suchi$^{1}$, Timothy Patten$^{1}$, David Fischinger$^{2}$, Markus Vincze$^{1}$
\thanks{$^{1}$M. Suchi, T. Patten, M. Vincze, Vision for Robotics Laboratory, Automation and Control Institute, TU Wien, 1040 Vienna, Austria. \{msu,tp,vm\}@acin.tuwien.ac.at}
\thanks{$^{2}$D. Fischinger, Aeolus Robotics, Inc., 1010 Vienna, Austria. david.fischinger@gmail.com}
\thanks{This work is partially supported by Aeolus Robotics, Inc., and the Austrian Science Foundation and CHIST-ERA, FWF I1856-N30, ALOOF, and WWTF ICT15-045, RALLI.}%
}

\begin{document}
\maketitle

\begin{abstract}
Developing robot perception systems for recognizing objects in the real-world requires computer vision algorithms to be carefully scrutinized 
with respect to the expected operating domain. This demands large quantities of ground truth data to rigorously evaluate the performance of algorithms. This paper presents the EasyLabel tool for easily acquiring high quality ground truth annotation of objects at the pixel-level in densely cluttered scenes. In a semi-automatic process, complex scenes are incrementally built and EasyLabel exploits depth change to extract precise object masks at each step. We use this tool to generate the Object Cluttered Indoor Dataset (OCID) that captures diverse settings of objects, background, context, sensor to scene distance, viewpoint angle and lighting conditions. OCID is used to perform a systematic comparison of existing object segmentation methods. The baseline comparison supports the need for pixel- and object-wise annotation to progress robot vision towards realistic applications. This insight reveals the usefulness of EasyLabel and OCID to better understand the challenges that robots face in the real-world.
\end{abstract}

\section{Introduction}
Knowledge about objects is crucial for robotic applications. 
Vision tasks such as object detection, recognition, and segmentation provide vital information to enable robots to act appropriately in user environments and conduct high-level tasks like cleaning, tidying and personal assistance.

Service and home applications challenge existing robotic vision systems in several ways. Most notable is the large number of occluded and unknown objects in cluttered scenes, combined with environmental factors such as illumination changes and reflections. Robots are typically prepared for operation in the real-world by learning rich models that leverage large datasets~\cite{firman-cvprw-2016}. Particularly RGB-D datasets, that provide spatial and color information, often exploited in robotic vision. Despite the considerable number and variety of existing datasets, the vast majority only supply bounding box annotations in the RGB image. A number of pixel-wise annotated datasets are available, however, at the cost of manually drawing object outlines, which can be both inaccurate and extremely time consuming. The problem can be alleviated by aligning known models~\cite{marion2017pipeline} but this does not generalize to arbitrary objects or scenes.

\begin{figure}[t]
	\vspace{1.3ex}
    \centering            
          \includegraphics[width=\columnwidth]{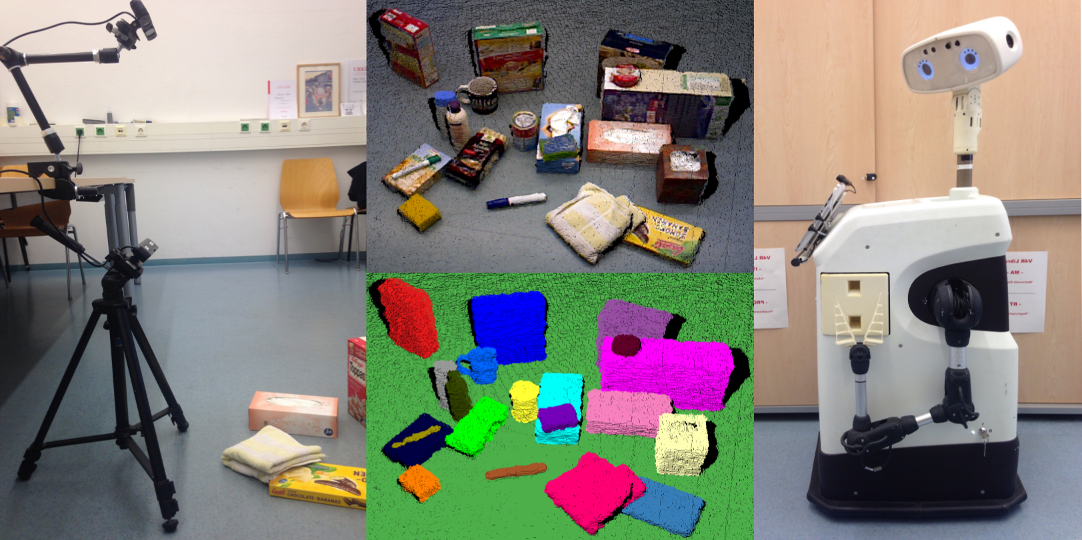} 
    \caption{EasyLabel - from recorded point clouds to pixel-wise labeled cluttered object datasets for robotic vision.}
    \label{easy}
    \vspace{-3ex}
\end{figure}

Current data annotation methods do not provide the necessary level of detail to deeply evaluate vision methods in densely cluttered and occluded scenes. To address this need we present EasyLabel~(cf. Figure~\ref{easy}), a semi-automatic ground truth pixel-wise annotation tool for RGB-D data. Ground truth data are generated during an incremental scene building process. Distance dependent depth change yields precise and unbiased annotation of the data for each object, which translates to pixel-wise object instance labels. The method does not require prior object knowledge, such as models, therefore, it can be used to label arbitrary objects including those that are non-rigid or deformable.

We use the EasyLabel tool to create the Object Clutter Indoor Dataset (OCID). The dataset is carefully structured to allow a statistical comparison of segmentation methods especially for increasing amount of clutter, distance from the sensor to the scene, background and viewpoint. The data varies in lighting condition, background and number of objects, and provides pixel-wise labeled data of 2346 scenes or 10,240 individual object masks. In addition to the labeled depth data, the annotation is projected to the RGB image to yield precise labeling in the 2D image enabling benchmarking of 2D image segmentation methods. Apart from segmentation, our dataset can be used to evaluate other vision methods such as object detection and classification with the automatically extracted bounding boxes and cropped images. We also present a thorough evaluation of segmentation methods based on color, depth and both modalities. The structure in the dataset enables a systematic exploration of the strengths and weaknesses of these methods.

Our contributions are as follows. (1) The EasyLabel tool to simply, yet effectively, generate high quality point- and pixel-wise ground truth annotation of objects for densely cluttered scenes without the need for prior models. (2) A new dataset that captures diverse settings of objects in clutter. (3)~Exploitation of the structured scene creation to meaningfully evaluate existing object segmentation methods and capture the relevant factors of the robotics domain.

\section{Related Work}
The increase of RGB-D datasets has contributed to algorithmic development in computer vision and robotics. Existing dataset cover a wide range of applications and needs~\cite{firman-cvprw-2016}. The NYU dataset~\cite{Silberman:ECCV12} has pixel-wise segmentation masks as well as corresponding class labels for structures and objects in a variety of indoor rooms. It is commonly used to benchmark and train vision algorithms, particularly those for semantic segmentation. The RGB-D Object Dataset~\cite{rod} is established for object classification and recognition by providing class labels of bounding boxes and cropped images. The Autonomous Robotic Indoor Dataset (ARID) extends~\cite{rod} by capturing images under real operating conditions. Such datasets have proven their worth for their respective domains, however, they do not necessarily generalize to other tasks. Object segmentation, for instance, performs fine-grained segmentation, as such a dataset for this purpose needs a high level of detail of both depth and color. Densely cluttered environments are particularly challenging to label, leading to limited volume of the current benchmark datasets~\cite{richtsfeld2012segmentation, ikkala2016benchmarking}.

Despite their established necessity, datasets are often hand annotated, which is time consuming and prone to human error. Modern tools, such as DeepExtremeCut~\cite{Man+18} that extend the pioneering work of GrabCut~\cite{Rother2004_grabcut}, significantly reduce human involvement to a small number of clicks. However, these tools are only usable for existing RGB images and do not facilitate depth data annotation. An interactive editor is introduced by Monica et al.~\cite{monica2017multi} for labeling 3D data, but this requires each instance to be labeled, therefore, it does not scale as the number of objects increases.

LabelFusion is a semi-automatic system for annotating 3D data in considerably less time~\cite{marion2017pipeline}. A human is required to roughly align a CAD model in a reconstruction, after which ICP is used to finely align the model in the reconstruction so that an object mask can be projected to each frame that was part of the reconstruction process. The method can generate vast quantities of labeled data. A limitation is that known CAD models are required and they must match the objects perfectly in the real-world, therefore, generalization to deformable or non-rigid objects is not possible. Furthermore, the data contains a large amount of redundancy because there is often a significant overlap between the frames that make up the reconstruction. An alternative approach, more similar to ours, is to use background subtraction to detect new objects in a scene~\cite{zeng2017APCdataset}. This method, however, has only been tested for single objects and involves a 2D pipeline that is sensitive to illumination fluctuations. Synthetic data generation is another direction for creating ground truth data for many object related vision tasks~\cite{McCormac:etal:ICCV2017}, but these lack the realism to capture all the characteristics of real sensor data.


Our method provides pixel-wise labels for real \mbox{RGB-D} data of cluttered scenes without requiring 3D models or manual annotation. EasyLabel only needs a human to record intermediate stages of an incrementally constructed scene. The procedure is simple, yet effective, for generating precise ground truth data to evaluate object segmentation algorithms.

\section{Semi-automatic Ground Truth Object Annotation with EasyLabel}\label{sec:method}
In this section we describe our simple strategy to semi-automatically label scenes with increasing object clutter. 
The process requires randomly selected objects to be placed one at a time. Data are captured from a statically mounted depth sensor at each stage of the building process until the final stage is reached. Each incremental recording consists of several frames of organized point clouds. The individual points clouds from one time step are accumulated into a single point cloud. This enables the data to be temporally smoothed using the method in~\cite{uckermann2012real} by calculating the moving average of each depth measurement over all frames. Our method does not discard data with huge depth difference, but adapts the moving average calculation by reseting the mean value when large jumps in depth occur. The output is dense point clouds which are needed for achieving high quality results in the labeling stage.

The main idea for generating labels for individual objects is to detect the difference in the scene between two consecutive stages. Using the accumulated data, precise depth changes can be calculated. Depth changes are known to be highly sensitive to distance~\cite{Halmetschlager2018}, therefore, a quadratic scaled depth change threshold is applied to separate new depth measurements. This enables the changes near to the sensor to be measured at a much smaller scale while discarding changes induced by the increasing noise of the depth sensor at larger distances. Once the depth change is computed, distance clustering is performed to gain the new labeling for the introduced objects, which results in a densely labeled template for the intermediate scene. The generated templates from each stage is used to look up the corresponding object labels in later stages. Human involvement in the labeling process is therefore only required in the recording phase.



EasyLabel is packaged as a set of simple programs for each task in the pipeline and will be publicly available together with OCID on our webpage\footnote{\url{https://www.acin.tuwien.ac.at/object-clutter-indoor-dataset/}}. Once the recordings are made, the tools are executed in a batch wise manner to produce the resulting labeled data. Our method operates purely on depth data, therefore, it is also suitable for sensors that do not provide RGB information. This, however, limits the types of objects which can be used for annotation. It is unsuitable for translucent, very shiny metallic, and objects smaller than the level of the sensor noise\footnote{A comprehensive survey of recent depth sensors and their noise characteristics is available in \cite{Halmetschlager2018}.}. The same procedure could be used with RGB input by detecting the color changes, however, this is highly sensitive to the illumination conditions. This approach is unusable in anything more than very controlled conditions. Our approach, however, can be easily used in any environment and we encourage its use to add more data to our preliminary set that we will describe next.

\section{Object Clutter Indoor Dataset}
This section describes the characteristics of the dataset that is created with the EasyLabel tool. Our main motivation is to evaluate object segmentation methods in cluttered scenes. With the labeling tools it is possible to produce annotated RGB-D data from no clutter to dense clutter out of the box.

\begin{figure}[t]
\begin{subfigure}[t]{\columnwidth}
     \centering
     \includegraphics[width=0.9\columnwidth]{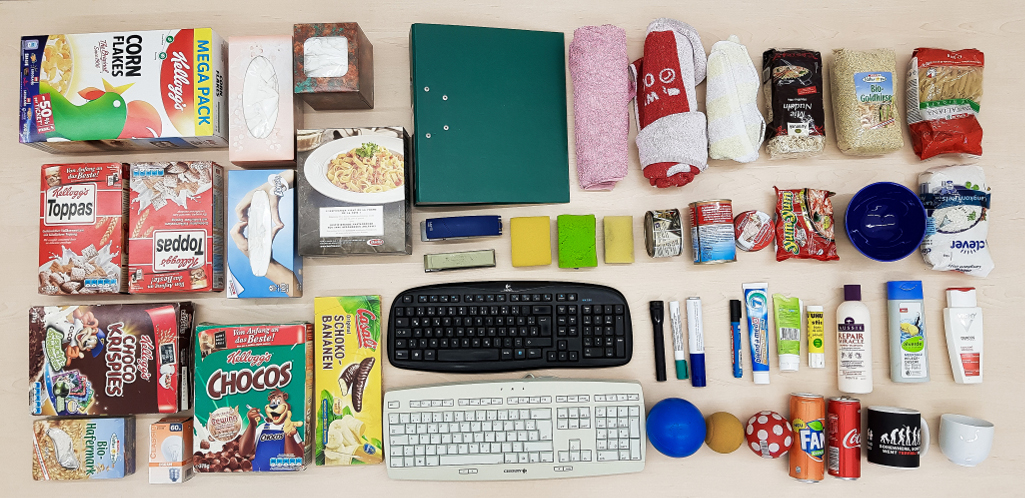}
     \caption{The common household instances from ARID.}
     \label{arid_objects}
\end{subfigure}
\begin{subfigure}[t]{\columnwidth}
     \centering
     \includegraphics[width=0.9\columnwidth]{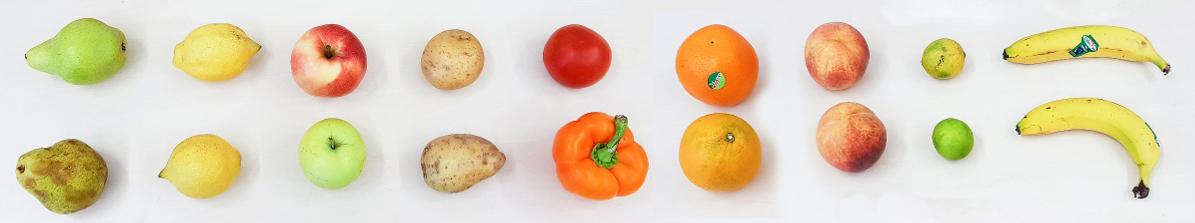}
     \caption{The organic object instances in ARID.}
     \label{arid_fruits_objects}
\end{subfigure}
\begin{subfigure}[t]{\columnwidth}
     \centering
     \includegraphics[width=0.9\columnwidth]{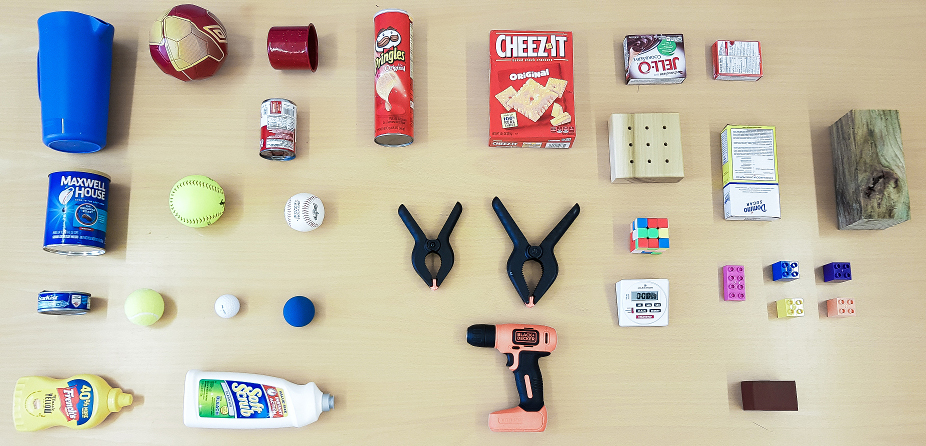}
     \caption{Instances from the YCB dataset.}
     \label{ycb_objects}
\end{subfigure}
\caption{Overview of the objects present in OCID. (a) A subset of the class instances from ARID. (b) The organic objects from ARID. (c) A subset of instances from the YCB dataset.}
\label{fig:objects_all}
\vspace{-2ex}
\end{figure}

An overview of the dataset structure is given in Table~\ref{dataset_overview}. OCID dataset comprises 96 fully built up cluttered scenes. The dataset uses 89 different objects (cf. Figure~\ref{fig:objects_all}) that are chosen representatives from the ARID classes and YCB dataset objects. The ARID classes are themselves a direct replication of the ROD classes. The ARID20 subset contains scenes including up to 20 objects from ARID. The ARID10 and YCB10 subsets include cluttered scenes with up to 10 objects from ARID and the YCB objects respectively. The scenes in each subset are composed of objects from only one of set at a time to maintain separation between datasets.


\begin{table*}[t]
\centering
\normalsize
\caption{Overview of OCID. The dataset is partitioned into subsets that comprise different number of scenes, categories and maximum number of objects. The number of labeled scenes and individual pixel-wise object instances are reported.}
\label{dataset_overview}
\begin{tabular}{ l l l l l l }
\textbf{Set} & \textbf{Scenes} & \textbf{Tags} & \textbf{Objects} & \textbf{Labels} & \textbf{Instances}\\
\hline
\text{ARID20} & \text{32} & \text{free, touching, stacked} & \text{20} & \text{1066} & \text{6720} \\ 
\text{ARID10} & \text{40} & \text{mixed, cuboid, curved, organic, non-organic} & \text{10} & \text{800} & \text{2200}  \\
\text{YCB10} & \text{24} & \text{mixed, cuboid, curved} & \text{10} & \text{480} & \text{1320} \\
\hline
\textbf{Total} & \textbf{96} & \text{} & \text{} & \textbf{2346} & \textbf{10240} \\
\end{tabular}
\vspace{-2ex}
\end{table*}

\begin{figure}[t]
     \centering
     \includegraphics[width=0.95\columnwidth]{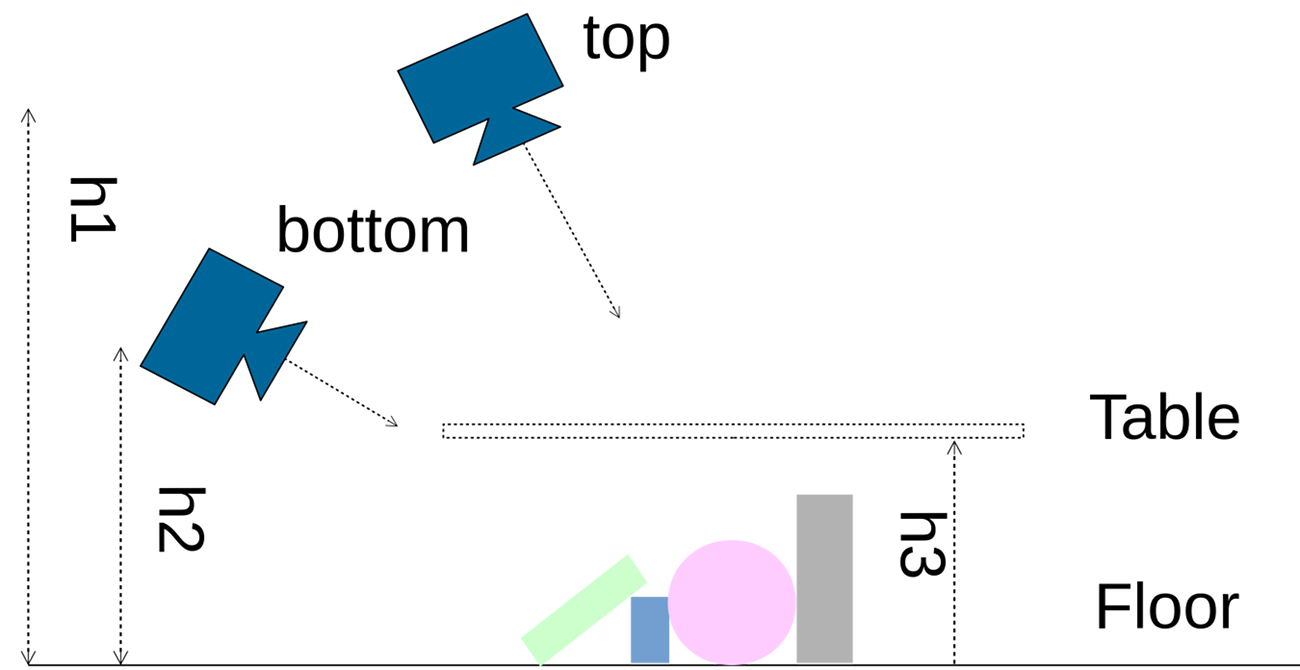}     
     \caption{Recording setup using two cameras mounted at different heights. Objects are placed on the floor or on a table (not present for floor scenes). Approximate distances for the table set up are: $h1\approx1.8$m, $h2\approx1.3$m, $h3\approx0.75$m. For the floor: $h1\approx1.0$m, $h2\approx0.6$m, $h3\approx0.0$m.}
     \label{camera_setup}
     \vspace{-2ex}
\end{figure}

Current depth sensors vary in the quality of their measurements for different distances, therefore, the dataset includes annotation of two distinct views of each scene as shown in Figure~\ref{camera_setup}. Each scene is simultaneously recorded by two ASUS-PRO cameras that are positioned at different heights. The chosen settings mimic configurations of existing robotic systems such as HOBBIT \cite{fischinger2015hobbitacare} and SQUIRREL~\cite{squirrel}. A common task for robots in home environments is picking up objects for users, which often involves objects on the floor or a table. We therefore included table and floor as separate categories of supporting planes for objects in the dataset.
As outlined in Section~\ref{sec:method}, multiple frames are recorded for temporal smoothing. We found that 20 frames was sufficient for good quality depth accumulation. Further scene variation includes different floor (plastic, wood, carpet) and table textures (wood, orange striped sheet, green patterned sheet).

To enable a systematic evaluation, OCID is neatly structured so that specific real-world factors can be individually assessed. The ARID20 dataset including up to 20 objects is structured into three clutter-level categories:
\begin{itemize}
\item Free: clearly separated (objects 1-9)
\item Touching: physically touching (objects 10-16)
\item Stacked: on top of each other (objects 17-20)
\end{itemize}
The YCB10 dataset has three shaped based categories:
\begin{itemize}
\item Cuboid: objects with sharp edges (e.g. cereal-boxes)
\item Curved: objects with smooth curved surfaces (e.g. ball)
\item Mixed: objects from both the cuboid and curved
\end{itemize}
In addition to the YCB10 categories, ARID10 includes organic objects (i.e. fruit and vegetables) and non-organic objects as categories.


\section{Evaluation}
This section presents the evaluation of common object segmentation methods. We describe the metrics used to evaluate the various methods and discuss the results. The evaluation is conducted on the ARID20 and YCB10 subsets.

\subsection{Metrics}
Comparing the segmentation output from an algorithm requires the labels of the ground truth to be matched with the labels of the segmentation image. The Hungarian algorithm is used to find the maximum matching overlaps between these two sets of labels. In preparation of the ground truth data, missing labels in the depth data are first filtered out.

The quality of the segmentation algorithms are measured by the unweighted mean of all objects in the scene. It is important to use an unweighted mean to avoid the bias imposed by the different size of the individual objects. For evaluation, the following metrics are used:
\begin{itemize}
\item{$p$, $r$, $f$}: precision, recall and F-score 
\item{$IoU$}: Intersection over Union 
\end{itemize}
The equivalent quantities that also exclude the background label and score are also used to more emphasize the objects in the scene. These are denoted with a subscript $nb$. Splitting the metrics between those that consider the background gives insight of how well not only objects are segmented but how well they are segmented from the supporting plane. Especially in the contexts that have colorful backgrounds, some segmentation methods are likely to perform worse. Background is considered to be anything that does not belong to the objects in the dataset, e.g. floor, table, wall, etc.

\subsection{Methods}
A large number of segmentation methods exists in the literature. We select methods that focus on objects, rather than, for example, semantics. This selection broadly represents the common schemes for the task and the different modalities available. The methods for this baseline comparison are summarized in Table~\ref{seg_methods}.

\emph{V4R}: The method developed in~\cite{potapova2014incremental} based on the approach of~\cite{richtsfeld2012segmentation}. This computes color and depth features for local patches, and uses a trained support vector machine to determine similarity scores between patches for grouping. The non-incremental version is used~\cite{v4r_library} and trained on OSD2.0.

\emph{LCCP}: The approach from~\cite{christoph2014object} that determines partitions in the input scene based on local convexity. This uses depth data and does not use a trained model. The publicly available implementation from the Point Cloud library (PCL)~\cite{pcl_library} is used with the original parameters in~\cite{christoph2014object}.

\emph{GCUT}: The graph-based image segmentation~\cite{Felzenszwalb2004_GraphBased} uses only color as input and requires no trained model. Boundaries between regions of an image are determined using a graph representation. The image is segmented by greedily making cuts in this graph. The original implementation is used~\cite{gcut_library} and best performance was achieved with parameters $\theta = 0.4$, $k = 500$ and the minimum cluster size of $500$.

\emph{SCUT}: SceneCut~\cite{pham2017scenecut} is a state-of-the-art approach combining object and semantic RGB-D segmentation using convolutional oriented boundaries (COB) and a hierarchical segmentation tree. Tests with RGB and RGB-D data (HHA encoding for depth) to compute ultrametric contour maps revealed better performance with RGB input. This is likely due to the higher degree of similarity between OCID and the PASCAL Context Dataset (PCD) used to train the COB network. Although depth provides additional information, the network is trained on the the NYU dataset, which focuses on semantics rather than object instances. Results are reported using the implementation from~\cite{scenecut_library} with the COB network trained on PCD with RGB input.

\begin{table}[t]
\centering
\normalsize
\caption{Characteristics of the used object segmentation implementations in the evaluation.}
\label{seg_methods}
\begin{tabular}{l l l l}
\textbf{Method} & \textbf{RGB} & \textbf{Depth} & \textbf{Learning Approach} \\
\hline
V4R & \checkmark & \checkmark & support vector machine \\
LCCP & - & - & - \\
GCUT & \checkmark & - & - \\
SCUT & \checkmark & \checkmark & deep learning \\ 
\end{tabular}
\vspace{-2ex}
\end{table}

\subsection{Quantitative analysis}
\begin{table*}[t]
\centering
\normalsize
\caption{Quantitative segmentation results for different methods on ARID20 and YCB10. Metrics reported are precision $p$, recall $r$, F-score $f$ and intersection over union $IoU$. The metrics with background label excluded are denoted with $nb$.}
\label{results_ARID20_20}
\begin{tabular}{l l l l l l l l l l}
\textbf{Subset} & \textbf{Method} & $\mathbf{p}$ & $\mathbf{r}$ & $\mathbf{f}$ & $\mathbf{IoU}$ & $\mathbf{p_{nb}}$ & $\mathbf{r_{nb}}$ & $\mathbf{f_{nb}}$ & $\mathbf{IoU_{nb}}$\\
\hline
\textbf{ARID20} & \text{GCUT} & \text{0.670} & \text{0.557} & \text{0.560} & \text{0.436} & \text{0.656} & \text{0.542} & \text{0.544} & \text{0.417} \\ 
                & \text{SCUT} & \text{0.735} & \text{0.705} & \text{0.692} & \text{0.603} & \text{0.724} & \text{0.696} & \text{0.680} & \text{0.590} \\
                & \text{V4R}  & \text{0.763} & \text{0.767} & \text{0.732} & \text{0.675} & \text{0.752} & \text{0.760} & \text{0.722} & \text{0.664} \\
                & \text{LCCP} & \textbf{0.925} & \textbf{0.825} & \textbf{0.848} & \textbf{0.787} & \textbf{0.922} & \textbf{0.822} & \textbf{0.844} & \textbf{0.783} \\ \hline

\textbf{YCB10} & \text{GCUT} & \text{0.623} & \text{0.543} & \text{0.513} & \text{0.385} & \text{0.587} & \text{0.553} & \text{0.511} & \text{0.379} \\ 
               & \text{SCUT} & \text{0.727} & \text{0.708} & \text{0.692} & \text{0.610} & \text{0.701} & \text{0.688} & \text{0.668} & \text{0.581} \\
               & \text{V4R}  & \text{0.770} & \text{0.787} & \text{0.762} & \text{0.720} & \text{0.748} & \text{0.767} & \text{0.739} & \text{0.694} \\
             & \text{LCCP} & \textbf{0.960} & \textbf{0.882} & \textbf{0.903} & \textbf{0.858} & \textbf{0.956} & \textbf{0.872} & \textbf{0.894} & \textbf{0.846} \\

\end{tabular}
\vspace{-2ex}
\end{table*}
The evaluation uses the ARID20 and YCB10 subsets of OCID. The first results in Table~\ref{results_ARID20_20} show overall average performance scores of all methods using complete scenes with 20 (ARID20) or 10 (YCB10) objects. GCUT as a pure RGB based method clearly performs worst. V4R shows better results than SCUT, but the difference is remarkably close on ARID20, given that SCUT is developed for scene and not object segmentation. LCCP, that heavily relies on depth data, shows the best performance. For V4R and LCCP the results are better for YCB10 since the number of objects is only 10. Surprisingly, this is not true for the methods that rely much more on RGB data. This indicates that the influence of the different backgrounds with textured carpet and table sheet strongly affect their performance.

\begin{figure*}[ht]
\centering
	\centering
    \begin{subfigure}[t]{0.32\textwidth}
       \centering
       \includegraphics[height=4.35cm,trim={2mm 0 1mm 0},clip]{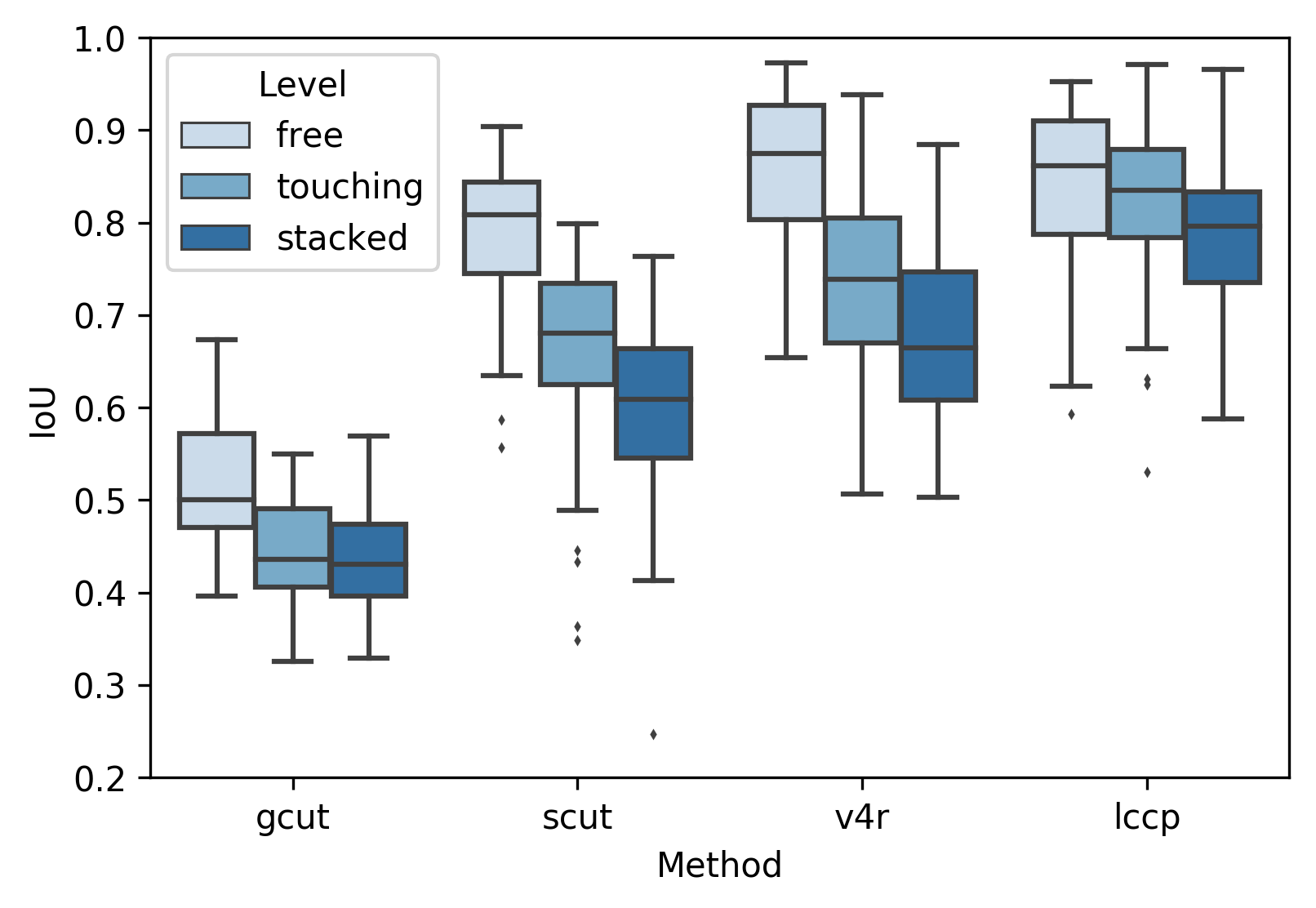}
       \caption{Clutter level.}
       \label{arid_clutter}
  	\end{subfigure}
    \hspace{3mm}
    \begin{subfigure}[t]{0.32\textwidth}
       \centering
       \includegraphics[height=4.3cm,trim={0mm 0 0mm 0},clip]{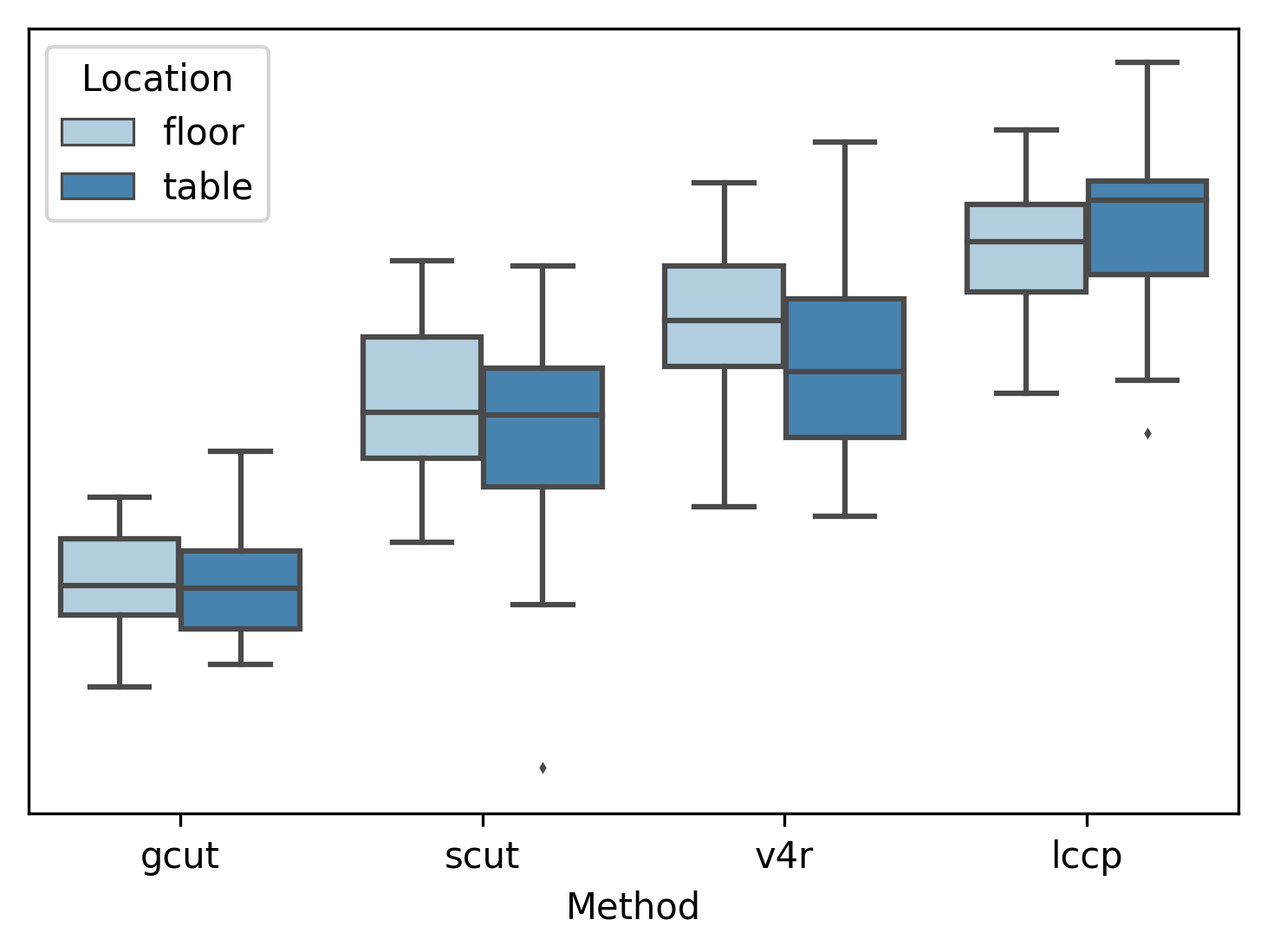}
       \caption{Supporting plane.}
       \label{arid_plane}
  	\end{subfigure}
    \hspace{-2mm}
    \begin{subfigure}[t]{0.32\textwidth}
       \centering
       \includegraphics[height=4.3cm,trim={0mm 0 0mm 0},clip]{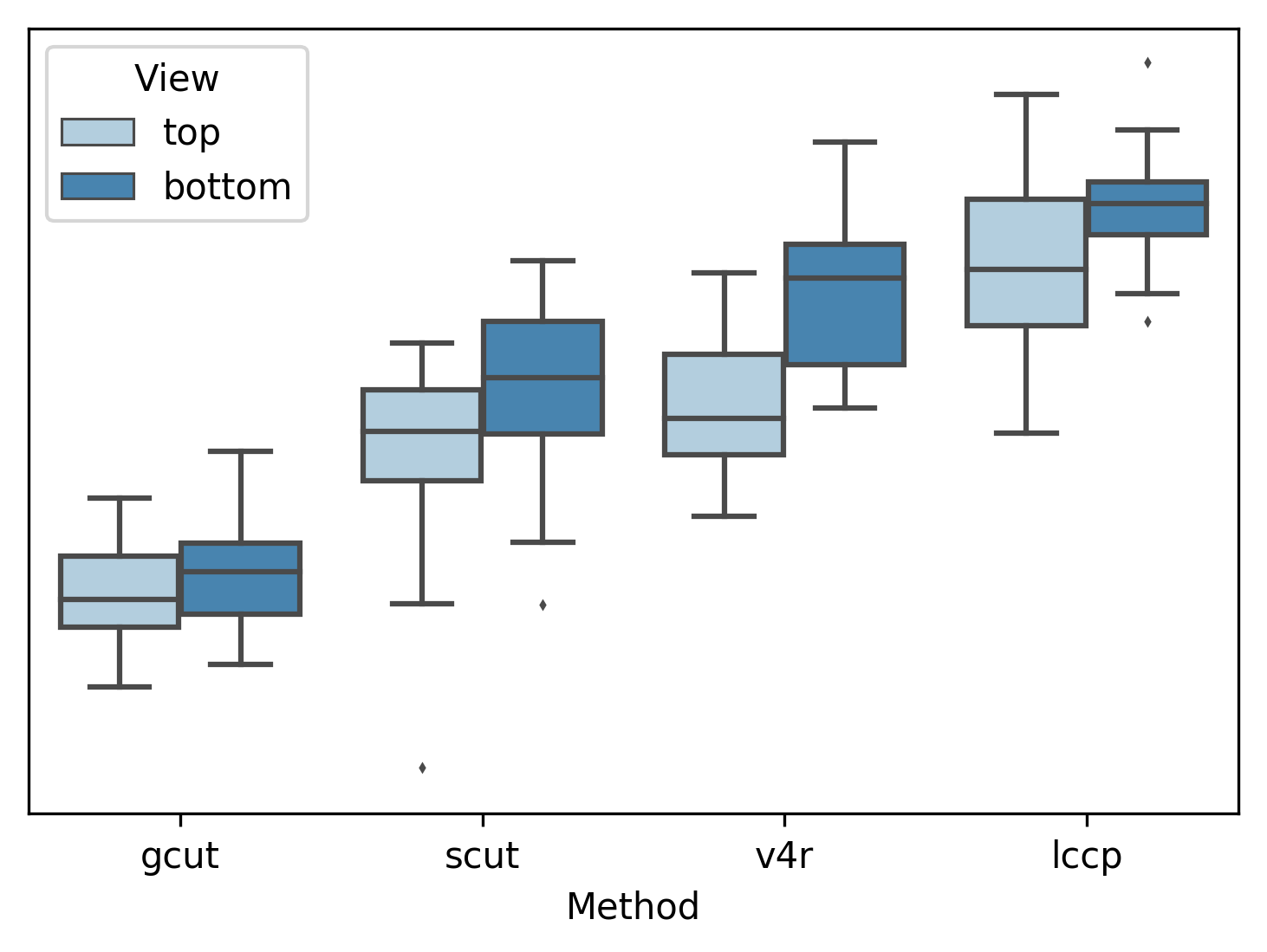}
       \caption{Camera viewpoint.}
       \label{arid_view}
  	\end{subfigure}
  \caption{Mean IoU scores for all segmentation methods of scenes with 20 objects from ARID20 subset. Comparison for different levels of clutter, supporting planes, and distances to objects.}
  \label{arid20_results}
  \vspace{-2ex}
\end{figure*}





Figure~\ref{arid_clutter} shows segmentation accuracy as clutter increases for ARID20. The figure reports the average IoU score of scenes containing the maximum amount of objects for each clutter level. As expected, all methods experience a performance decrease. LCCP is affected the least by the amount of objects and clutter. The performance of GCUT already declines when objects begin to touch, while the remaining methods maintain higher scores even for stacked objects. 

The influence of the supporting plane is visualized in Figure~\ref{arid_plane}, but it shows little differences because the background texture does not vary a great deal in ARID20. In contrast, the distance to the sensor plays a more significant role as illustrated in Figure~\ref{arid_view}. All methods performed better when the sensor is placed nearer to the objects. This is particularly prominent for SCUT and V4R, whereas for GCUT and LCCP only a slight performance increase is present.

\begin{figure*}[ht]
\centering
	\centering
    \begin{subfigure}[t]{0.32\textwidth}
       \centering
       \includegraphics[height=4.35cm,trim={2mm 0 1mm 0},clip]{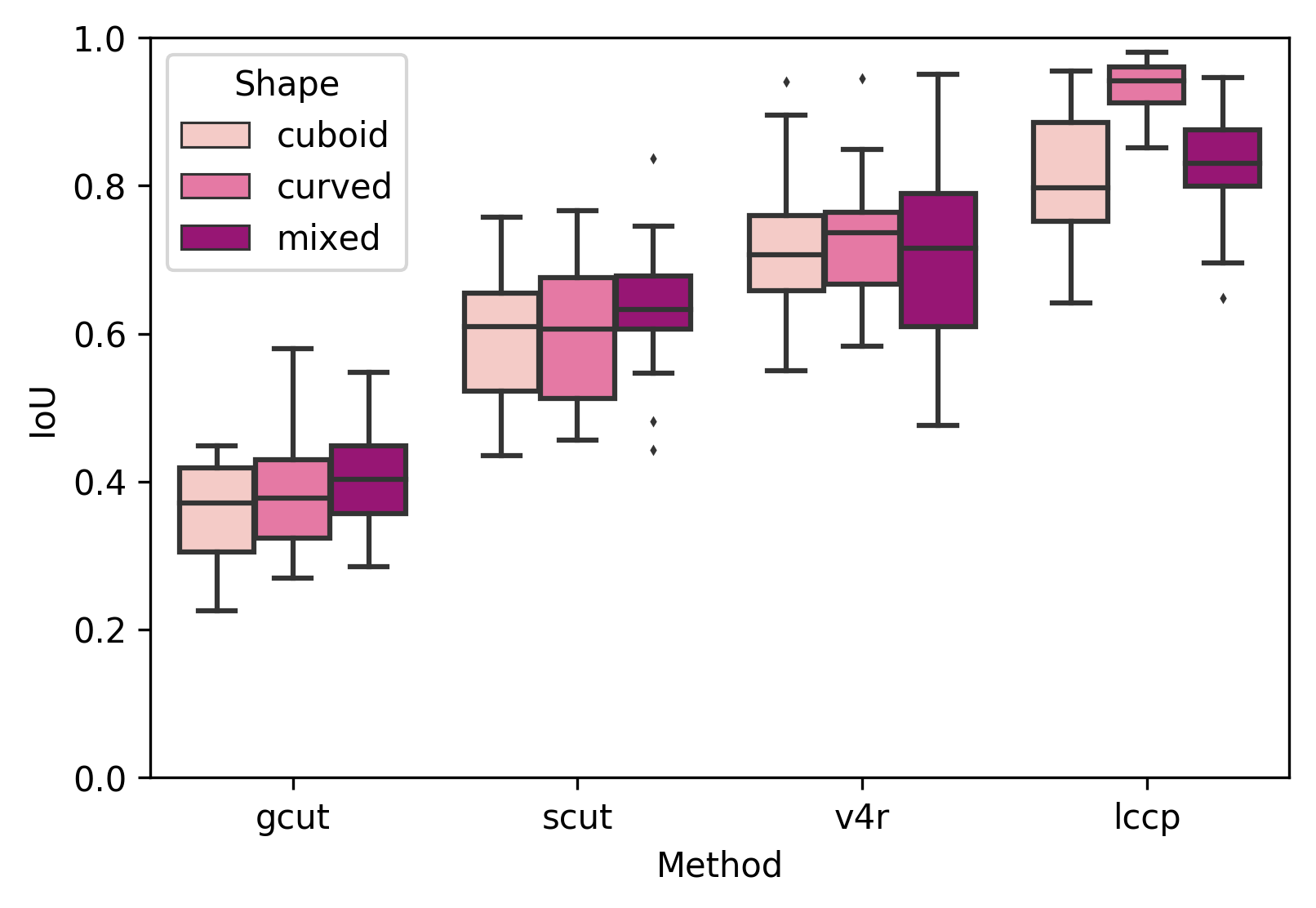}
       \caption{Shape.}
       \label{ycb_shape}
  	\end{subfigure}
    \hspace{3mm}
    \begin{subfigure}[t]{0.32\textwidth}
       \centering
       \includegraphics[height=4.3cm,trim={0mm 0mm 0mm 0},clip]{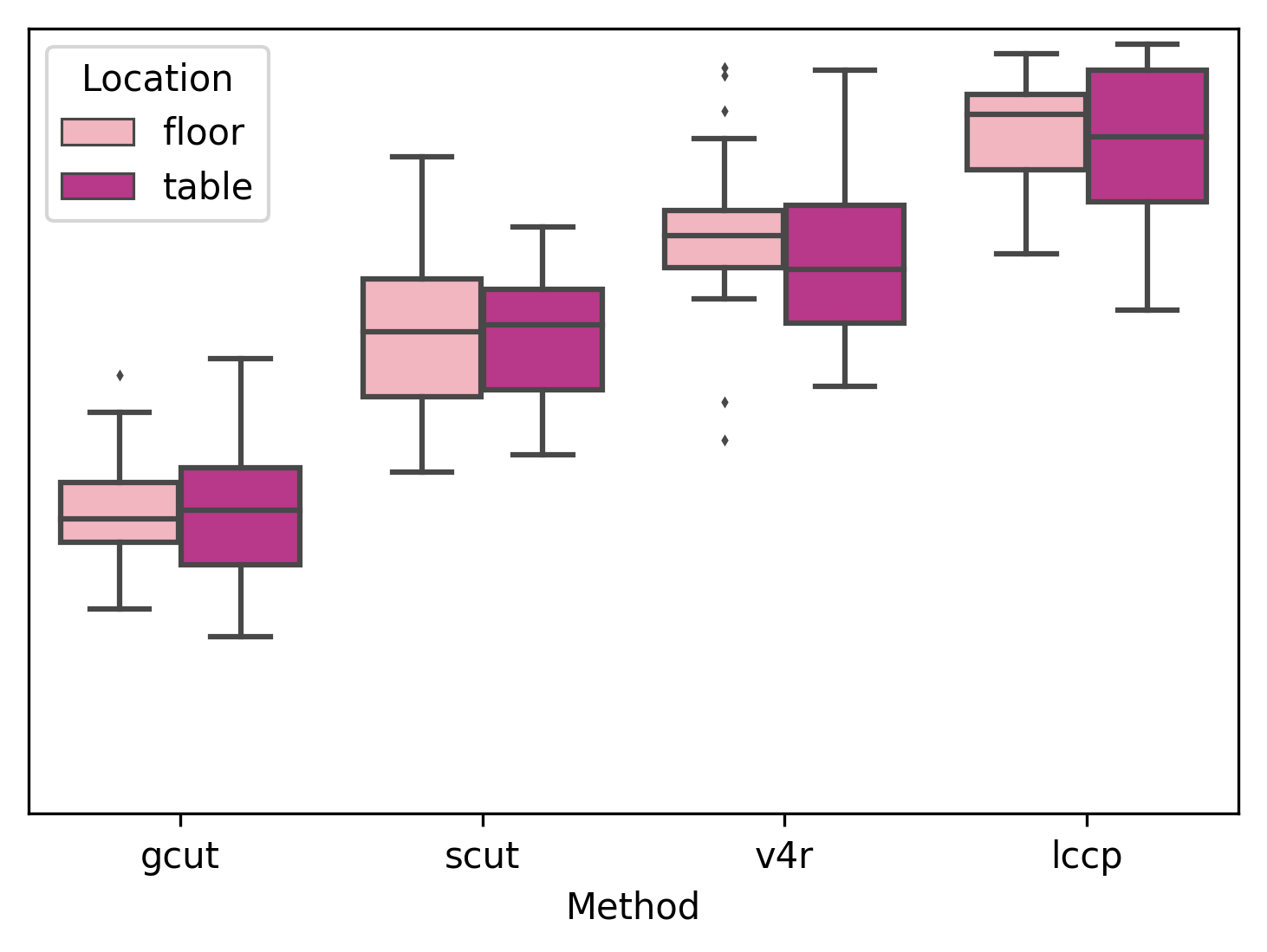}
       \caption{Supporting plane.}
       \label{ycb_plane}
  	\end{subfigure}
    \hspace{-2mm}
    \begin{subfigure}[t]{0.32\textwidth}
       \centering
       \includegraphics[height=4.3cm,trim={0mm 0mm 0mm 0},clip]{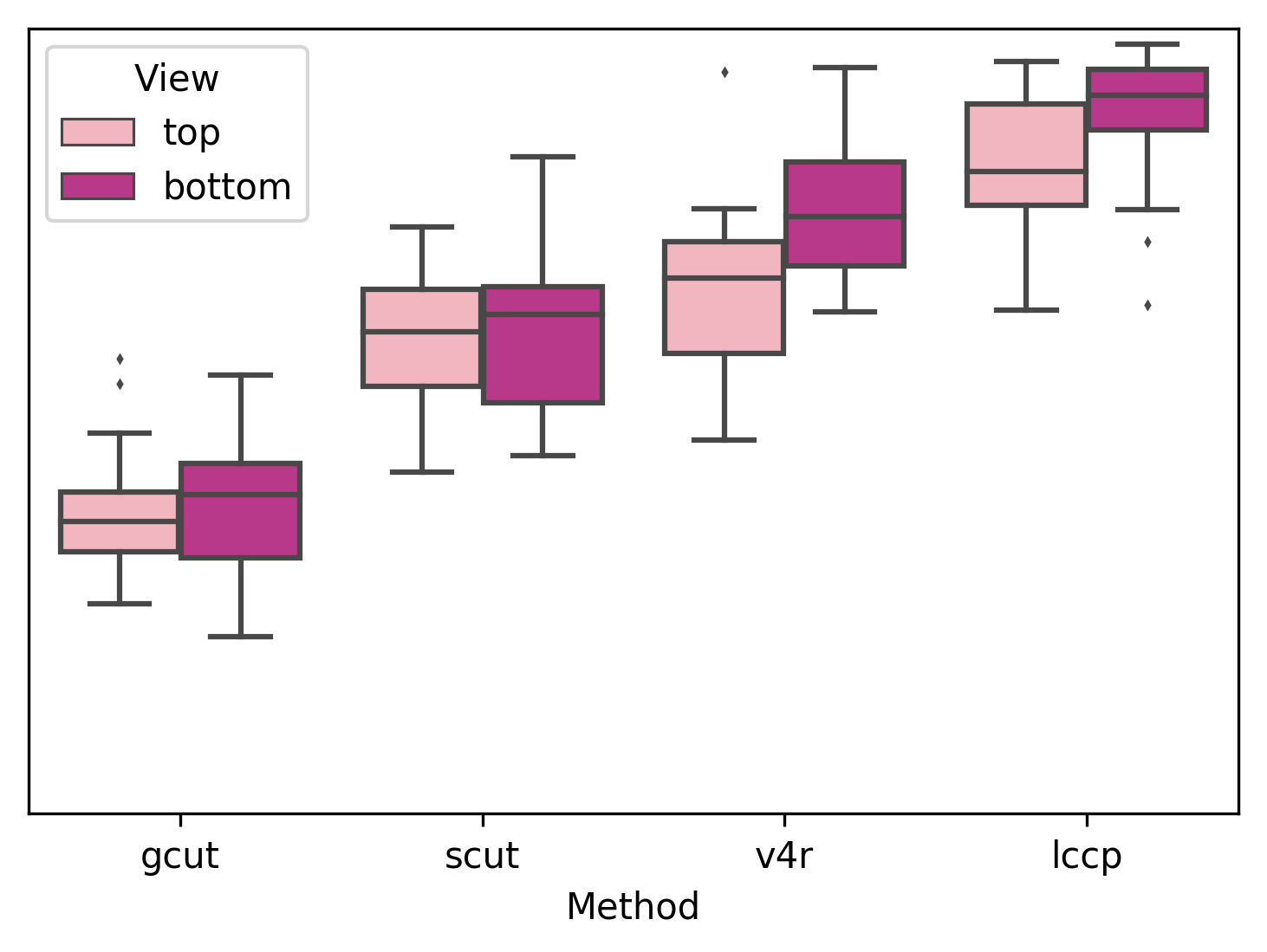}
       \caption{Camera viewpoint.}
       \label{ycb_view}
  	\end{subfigure}
  \caption{Mean IoU scores for all segmentation methods of scenes with 10 objects from YCB10 subset. Comparison for different shaped objects, supporting planes, and distances to objects.}
  \label{ycb10_results}
  \vspace{-2ex}
\end{figure*}



An analysis of the shape categories in YCB10 is presented in Figure~\ref{ycb_shape}. Except for the curved shapes most methods performed similar for all categories. LCCP is particularly good with the these objects, which is sensible since the method is based on local curvature. In agreement with the results on ARID20, GCUT, V4R and LCCP show a similar trend with the different sensor to object distances. SCUT, on the other hand, shows less significant change.

The influence of different supporting planes is not very significant for all methods as shown in Figure~\ref{ycb_plane}. The interesting observation, however, is that SCUT and GCUT could not improve on the YCB10 scenes, revealing the source of the overall performance drop from ARID20 to YCB10 (cf. Table~\ref{results_ARID20_20}). These methods struggle for consistency, which is caused by the different and extreme floor and table textures used in the dataset.

\begin{figure*}[tb]
     \centering
     \includegraphics[width=0.75\linewidth]{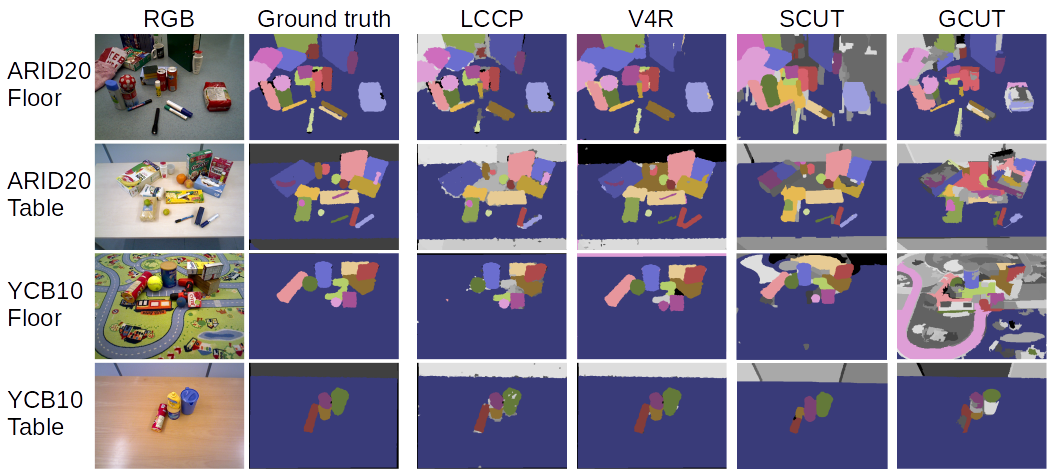}
     \caption{Ground truth labels from EasyLabel and qualitative results of segmentation methods for OCID scenes. Colored patches show matching overlaps with corresponding ground truth labels. Grey patches have no matching.}
     \label{result_vis}
     \vspace{-2ex}
\end{figure*}
Figure~\ref{result_vis} presents a qualitative evaluation of the tested methods on example scenes in OCID. It is clearly visible that methods favoring depth over RGB information produce more accurate segmentation. The underwhelming results of the methods relying on color is highlighted in the scene with an extremely textured background. The details of the carpet are segmented together with objects. The visualization also provides a good impression of the high quality pixel-wise ground truth labeling retrieved using Easylabel.

\section{EasyLabel and OCID for Other Vision Tasks}

\begin{figure}[t]
\begin{subfigure}[t]{0.49\columnwidth}
     \centering
     \includegraphics[width=\columnwidth]{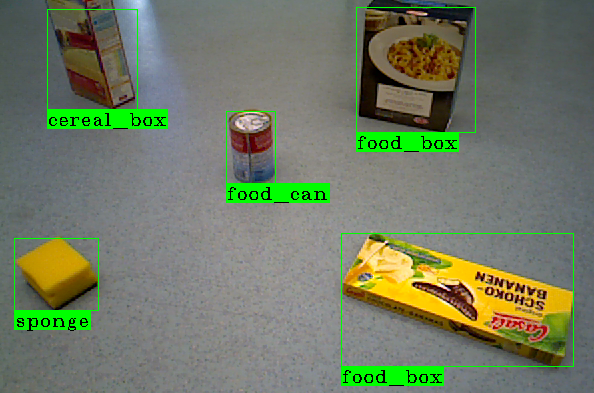}
     \caption{Intermediate frame (5 objects).}
     \label{class_5}
\end{subfigure}
\begin{subfigure}[t]{0.49\columnwidth}
     \centering
     \includegraphics[width=\columnwidth]{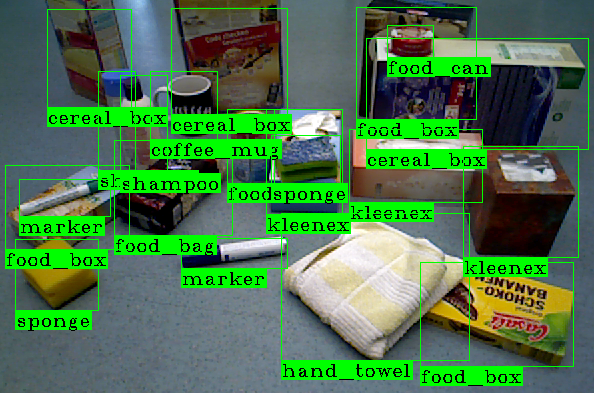}
     \caption{Final frame (20 objects).}
     \label{class_20}
\end{subfigure}
\begin{subfigure}[t]{\columnwidth}
     \centering
     \includegraphics[width=\columnwidth]{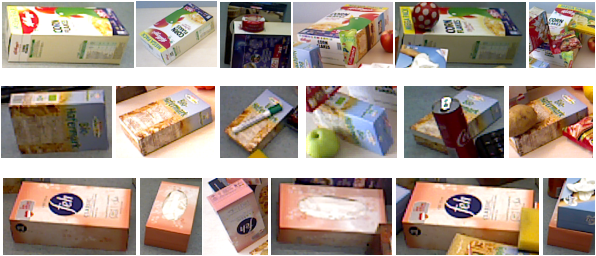}
     \caption{Examples of cropped images for three different instances in ARID20, arranged with increasing clutter from left to right.}
     \label{sampl_crops}
\end{subfigure}
\caption{Automatically extracted ground truth bounding boxes and cropped images of object instances with EasyLabel.}
\label{bboxes}
\vspace{-2ex}
\end{figure}

The primary motivation behind the EasyLabel tool is to generate detailed pixel-wise labels of cluttered scenes to evaluate and develop segmentation methods. However, the pixel-wise labeling can be automatically converted to other ground truth data formats and therefore used to evaluate many other vision tasks.

\emph{Object detection/classification} is enabled through automatically generated 2D bounding boxes and cropped images as shown in Figure~\ref{bboxes}. While class labels must be hand annotated, this is only necessary for the final frame. Labels propagate through the incremental frames to generate a large quantity of individual labeled instances in the sequence.

\begin{figure}[t]
\begin{subfigure}[t]{0.49\columnwidth}
     \centering
     \includegraphics[width=\columnwidth]{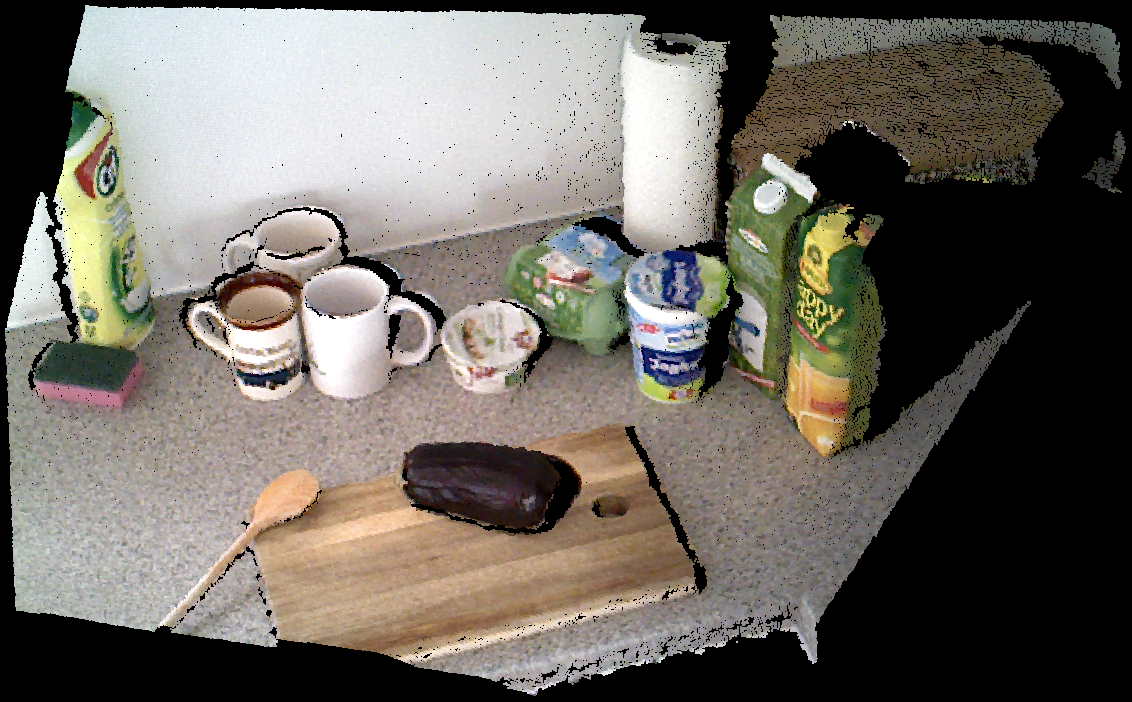}
     \caption{Final point cloud of incrementally constructed scene.}
     \label{rgb_kitchen}
\end{subfigure}
\begin{subfigure}[t]{0.49\columnwidth}
     \centering
     \includegraphics[width=\columnwidth]{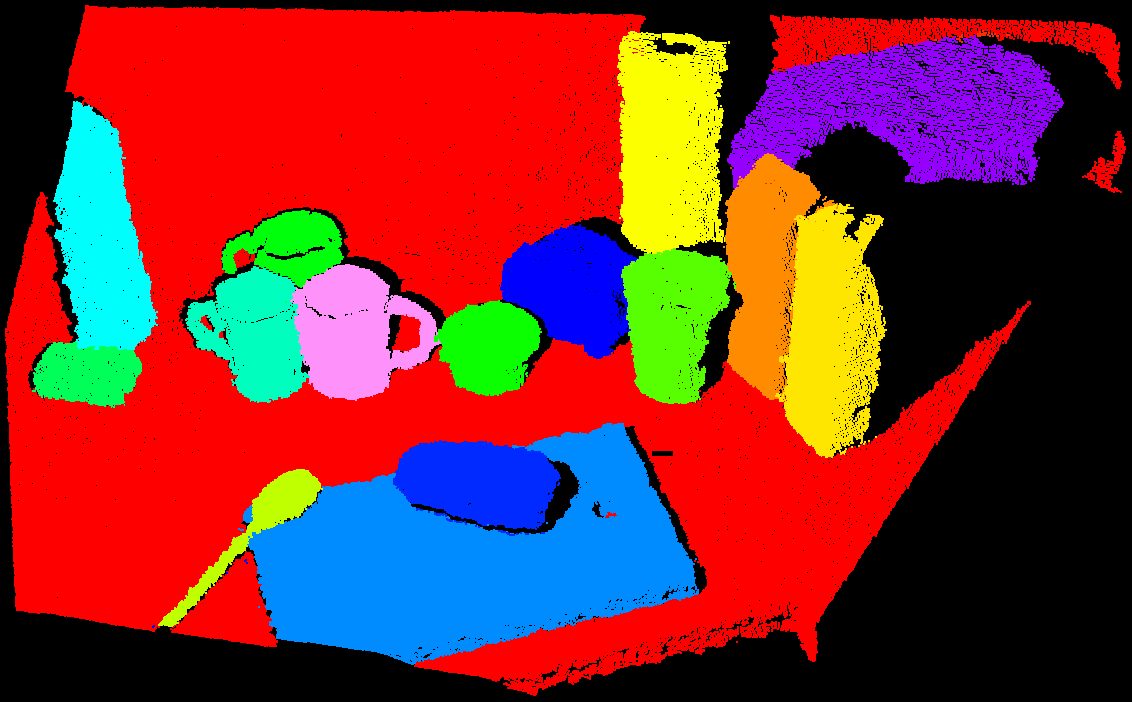}
     \caption{Ground truth pixel-wise annotation of objects.}
     \label{label_kitchen}
\end{subfigure}
\caption{Example use of EasyLabel to annotate a real-world indoor scene (kitchen).}
\label{kitchen}
\vspace{-2ex}
\end{figure}

\emph{Real-world setups} are easily annotated using the same principles. While OCID is created specifically for cluttered scenes of objects, any non-static set of objects in real scenarios can be labeled as shown in Figure~\ref{kitchen}.

\emph{Interactive segmentation} requires ground truth labels for consecutive frames that may differ by small differences. Our approach can generate data for this purpose as has been demonstrated in previous work~\cite{Patten2018} that used a preliminary version of EasyLabel to generate ground truth object labels.

\section{Discussion And Conclusion}

This work introduces a simple, fast and efficient method for retrieving pixel-wise annotated RGB-D data on an object level. We propose an approach where the complexity of a scene is increased by placing one object at a time. Recordings from each step are automatically processed by exploiting spatial shifts in the depth data to generate the ground truth annotation. The methods requires no object models, only a single depth sensor, to reduce human time for labeling to simple recording and qualitative inspection.

Our procedure enables large ground truth data to be created in a controlled way. Our dataset, the Object Clutter Indoor Dataset, consists of different objects, backgrounds and lighting conditions to provide a large variety for evaluation purposes and capture aspects relevant to robotic systems. OCID enables us to provide answers about the performance of object segmentation methods with respect to the influence of clutter, distance to the scene, supporting structures and object shape. Results show that methods relying largely on RGB information are particularly challenged in this dataset. Scenes that are highly textured, or with objects and backgrounds of similar color, favor geometry-based methods.

Future work will address the imbalance of OCID and extend it with more demanding content for geometry based methods. We will also extend the volume of data by propagating labels from EasyLabel to scene reconstruction, e.g. in combination with LabelFusion~\cite{marion2017pipeline}, to label templates for annotating individual frames for many viewpoints. This will enable us to produce data on the scale that is necessary to train deep learning approaches. We believe that our tool and datasets will be immensely valuable for developing various vision methods, through evaluation and training, to extend their capabilities in cluttered scenes.
\balance
\bibliographystyle{IEEEtran}
\bibliography{Bibliography}

\end{document}